\documentclass[conference,a4paper]{IEEEtran}
\IEEEoverridecommandlockouts

\usepackage[hidelinks]{hyperref}
\usepackage[cmex10]{amsmath}
\usepackage{amssymb,amsfonts}
\usepackage{dblfloatfix}

\usepackage[ruled,vlined]{algorithm2e}
\usepackage{graphicx}
\graphicspath{{Figures/PDF/}{Figures/PNG/}}

\usepackage{booktabs}
\usepackage{siunitx}
\usepackage[numbers,compress]{natbib}
\usepackage{texnames}
\usepackage{bm,bbm}
\usepackage{orcidlink}
\usepackage{graphicx}
\usepackage{multirow}
\usepackage{tablefootnote}
\usepackage{adjustbox}
\usepackage{array} 

\begin{document}

\title{SAR-W-MixMAE: SAR Foundation Model\\Training Using Backscatter Power Weighting 
\thanks{*Authors contibuted to this work equally.}
}

\author{	\IEEEauthorblockN{Ali Caglayan*\orcidlink{0000-0002-3408-8659}}
	\IEEEauthorblockA{\textit{National Institute of Advanced}\\
        \textit{Industrial Science and Technology}\\
		135-0064 Tokyo, Japan\\
		ali.caglayan@aist.go.jp}
	\and
	\IEEEauthorblockN{Nevrez Imamoglu*\orcidlink{0000-0002-2661-599X}}
	\IEEEauthorblockA{\textit{National Institute of Advanced}\\
        \textit{Industrial Science and Technology}\\
		135-0064 Tokyo, Japan\\
		nevrez.imamoglu@aist.go.jp}
	\and
	\IEEEauthorblockN{Toru Kouyama\orcidlink{0000-0002-1060-3986}}
	\IEEEauthorblockA{\textit{National Institute of Advanced}\\
        \textit{Industrial Science and Technology}\\
		135-0064 Tokyo, Japan\\
		t.kouyama@aist.go.jp}
}

\maketitle
\begin{abstract}
    Foundation model approaches such as masked auto-encoders (MAE) or its variations are now being successfully applied to satellite imagery. Most of the ongoing technical validation of foundation models have been applied to optical images like RGB or multi-spectral images. Due to difficulty in semantic labeling to create datasets and higher noise content with respect to optical images, Synthetic Aperture Radar (SAR) data has not been explored a lot in the field for foundation models. Therefore, in this work as a pre-training approach, we explored masked auto-encoder, specifically MixMAE \cite{liu2023mixmae} on Sentinel-1 SAR images and its impact on SAR image classification tasks. Moreover, we proposed to use the physical characteristic of SAR data for applying weighting parameter on the auto-encoder training loss (MSE) to reduce the effect of speckle noise and very high values on the SAR images. Proposed SAR intensity-based weighting of the reconstruction loss demonstrates promising results both on SAR pre-training and downstream tasks specifically on flood detection compared with the baseline model.
\end{abstract}


\section{Introduction}
\label{sec:intro}
Foundation models have advanced the field of computer vision \cite{liu2023mixmae, he2022masked, kirillov2023segment, zhang2022dino}, enabling large-scale pretraining on unlabeled data. These models, such as masked autoencoders (MAEs) \cite{he2022masked}, use self-supervised learning to make pretraining more effective and minimize the use of labeled datasets. In remote sensing field, important progress has been made with foundation models, mostly applied to optical images like RGB \cite{wang2022advancing, ayush2021geography} and multispectral data \cite{hong2024spectralgpt, noman2024rethinking, li2024s2mae}. However, synthetic aperture radar (SAR) data, despite its role in environmental monitoring \cite{brisco2020hybrid}, disaster response (\textit{e.g.} flood detection \cite{amitrano2024flood}), and land-cover mapping \cite{waske2009classifier}, remains less explored in foundation models. Studies such as SpectralGPT \cite{hong2024spectralgpt} and CROMA \cite{fuller2024croma} have shown the promise of leveraging domain-specific priors to improve feature representation in spectral and multi-modal remote sensing data. Moreover, methods like S2MAE \cite{li2024s2mae} have emphasized the importance of spatial-spectral pretraining, indicating the need for domain-adaptive solutions. 

\begin{figure*}[t!]
	\centering
	\includegraphics[width = 0.85\textwidth]{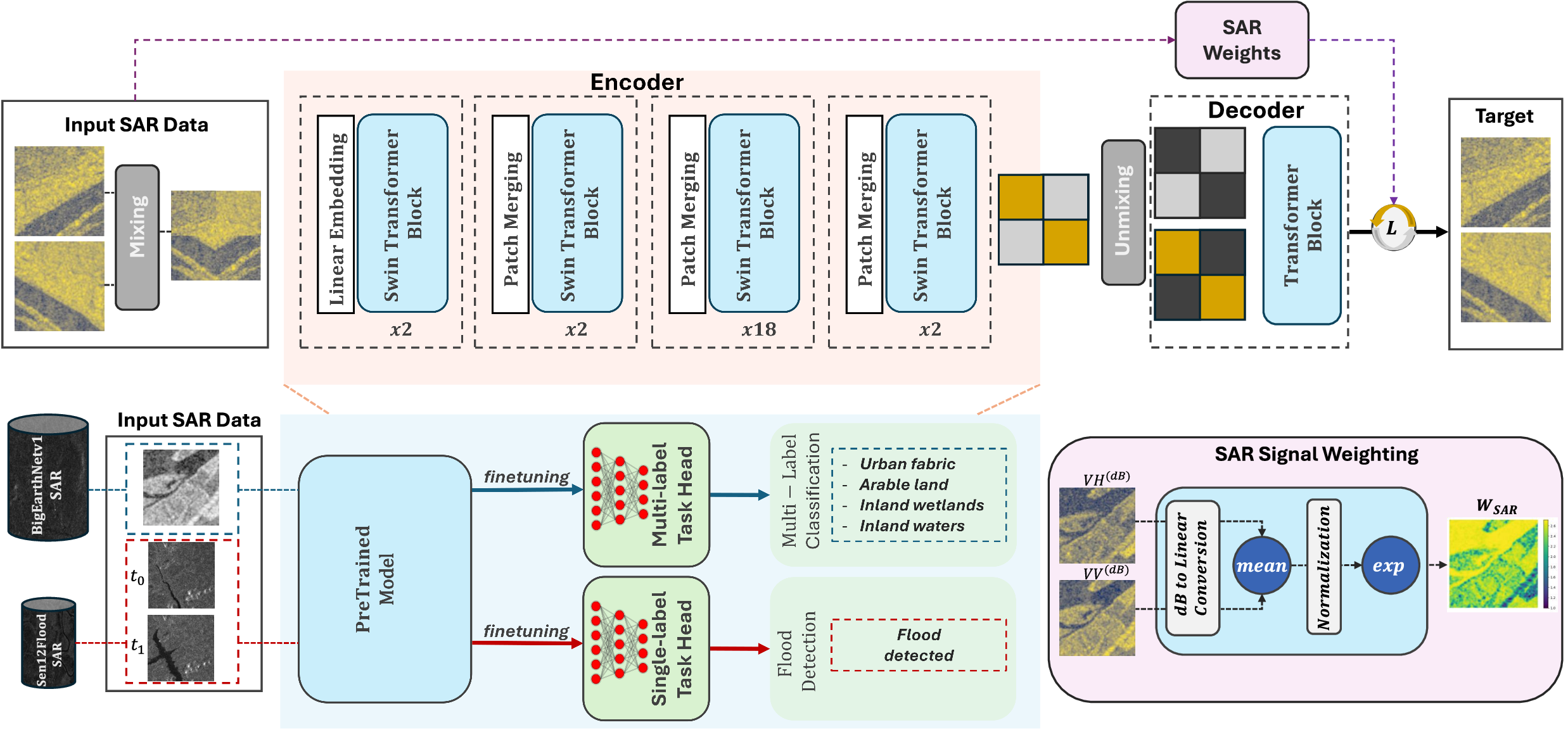}
	\centering
    \caption{Overview of the proposed SAR-W-MixMAE model. Sentinel-1 SAR data with VH and VV channels is processed through a mixing module to combine patches from two inputs, followed by a hierarchical encoder leveraging Swin Transformer blocks and patch merging. SAR-specific pixel-wise weights are incorporated into the reconstruction loss during the decoding phase to enhance robustness against SAR noise. The pretrained model is further fine-tuned for downstream tasks such as multi-label classification and flood detection.}\label{fig:method}
\end{figure*}

SAR data relying on radar imaging techniques that enable observations in all weather conditions, day or night, differs significantly from optical sensor based imaging, such as RGB images, due to its unique properties \cite{moreira2013sartutorial, nasa_sar_basics, rs15030839, geng2017deep, sumbul2019bigearthnet}. \textit{(i)} SAR data is characterized by multiplicative speckle noise. Unlike additive noise, its multiplicative character prevents reduction of speckle just by increasing the transmit signal power \cite{moreira2013sartutorial}, complicating direct application of traditional image modeling techniques used in optical imagery. \textit{(ii)} SAR intensities have substantial variability, often requiring specific transformations for effective feature extraction. \textit{(iii)} SAR images (\textit{i.e.} VV, VH, HH, and HV) encode how electromagnetic waves are transmitted and received (vertical or horizantal) through radar antennas during image generation \cite{moreira2013sartutorial, rs15030839}. 
SAR imagery obtained through different polarizations and wavelengths provide complementary information that can be critical for earth observation tasks including land use, urban area detection, snow/ice detection, forest monitoring, and water/flood analysis \cite{moreira2013sartutorial,nasa_sar_basics}. For instance, brighter areas in SAR images often indicate rough surfaces, man-made structures, or similar features; on the other hand, calm water areas generally having low backscatter power, resulting in darker regions in SAR images \cite{nasa_sar_basics, moreira2013sartutorial, esa_sentinel1}. Therefore, these characteristics make SAR data invaluable for task-specific earth observation analysis. MixMAE \cite{liu2023mixmae} has demonstrated the efficiency of mixing-based masked autoencoders for RGB data in computer vision, but its direct application to SAR images remains suboptimal due to the unique noise and signal characteristics of SAR. Considering these challenges and the characteristics of SAR images, in this work:

\begin{itemize}
    \item First, we explore the use of masked auto-encoder, specifically MixMAE \cite{liu2023mixmae}, for pre-training on Sentinel-1 \footnote{\url{https://www.esa.int/Applications/Observing_the_Earth/Copernicus/Sentinel-1}} \cite{esa_sentinel1} dual-polarization SAR data (VH and VV) in decibel (dB) scale.
    \item We propose SAR-W-MixMAE, which incorporates VH and VV polarimetric SAR channels in linear scale for pixel-wise weighting during pre-training (see. Fig.\ref{fig:method}). This approach makes the reconstruction loss less sensitive to the varying reliability of SAR backscattered signal power with speckle while giving more importance to optimization based on areas with relatively lower-signal power areas such as water on the sample data. 
    \item We evaluate SAR-W-MixMAE pretrained model on two downstream tasks: i) multi-label SAR image classification to verify if the weighting in favor of low-scatter power regions have adverse affect on  multi category classification task, and ii) SAR flood detection demonstrating that SAR-W-MixMAE improves baseline MixMAE model with significant margins on flood detection. 
\end{itemize}

\section{Proposed Method}
\label{sec:method}

Before introducing the proposed SAR-W-MixMAE, we briefly explain the baseline MixMAE\footnote{\url{https://github.com/Sense-X/MixMIM}}  method \cite{liu2023mixmae}. MixMAE addresses key limitations of masked image modeling \cite{he2022masked}, which suffers from training inefficiency due to large portions of image patches being masked and replaced with a special symbol [MASK]. These masked regions contribute little to the pretraining process over long epochs, and the special symbols create inconsistencies between pretraining and fine-tuning since they do not appear during the fine-tuning stage \cite{liu2023mixmae}. To overcome these challenges, MixMAE introduces a mixing strategy during pretraining. Instead of masking patches and replacing them with placeholders, MixMAE takes two random training images, \(x_1\) and \(x_2\), and creates a mixed image by filling the masked regions of one image with corresponding regions from the other. This mixed image is used as input for the reconstruction task, where the model performs a dual reconstruction task by “unmixing” the input to reconstruct the original images \(x_1\) and \(x_2\). By simultaneously learning to reconstruct two original images from their mixed representation, MixMAE improves pretraining efficiency and alignment with fine-tuning.

Building on MixMAE, the proposed SAR-W-MixMAE adapts this framework for Sentinel-1 SAR data, addressing the unique challenges posed by SAR imagery, such as its inherent noise and signal characteristics \cite{ding2016convolutional, geng2017deep}. 
To address these challenges, SAR-W-MixMAE introduces pixel-wise data weighting into the MixMAE pre-training, making the reconstruction loss sensitive to the varying reliability of SAR backscattered signal power with speckle. The primary goal is to modulate reconstruction loss to handle challenges of SAR images during training and to create more robust representation during the pretraining. To do so, we compute a weight matrix $\text{W}_\text{SAR}$ from input data itself for each pixel as follows:

\begin{equation}
    \text{W}_\text{SAR} = \exp \left( 1.0 - \text{norm}\left( \sigma^0_{\text{avr}}\text{(linear)}  \right) \right)
\label{eq:weight_sar_w_mixmae}
\end{equation}

In Eq.\ref{eq:weight_sar_w_mixmae}, $\sigma^0_{\text{avr}}\text{(linear)}$ refers to the linear scale averege of VH and VV polarimetric SAR channels, which is also scaled to the range [0,1] with the \(\text{norm}(\cdot)\) as min-max normalization.  The calculation of $\sigma^0_{\text{avr}}\text{(linear)}$ is given in Eq.\ref{eq:VH_VV_avr_linear_scale}, where original decibel (dB) scale SAR channels ($\text{VH}^\text{(dB)}$ and $\text{VV}^\text{(dB)}$) are converted from dB to linear scale signal power \cite{esa_sentinel1,moreira2013sartutorial} as in  Eq.\ref{eq:VH_VV_linear_scale}.


\begin{equation}
\sigma^0_{\text{avr}}\text{(linear)} = \frac{1}{2} \sum_{p \in \{\text{VH}, \text{VV}\}} \sigma^0_{p}\text{(linear)}
\label{eq:VH_VV_avr_linear_scale}
\end{equation}


\begin{equation}
    \sigma^0_{p}\text{(linear)} = 10^{\frac{\sigma^0_{p}\text{(dB)}}{10}}, \quad p \in \{\text{VH}, \text{VV}\}
    \label{eq:VH_VV_linear_scale}
\end{equation}

The computed weight matrix $\text{W}_\text{SAR}$ yields higher weights for lower backscatter signal power (relatively darker regions on SAR images) and lower weights on pixels with stronger backscatter power at linear scale. It is utilized to emphasizes homogeneous or uniform relatively low backscatter power pixels and less noisy regions more on the gradients during backpropagation for parameter update process, which is aimed to provide more robust learning from challenging areas and speckle. The reconstruction loss in \cite{liu2023mixmae} is modified to incorporate the proposed pixel-wise weights ( $W_{SAR}$ see Eq.\ref{eq:weight_sar_w_mixmae} and Fig.\ref{fig:method}) as follows:

\begin{equation}
\begin{adjustbox}{width=0.9\columnwidth}
    {$
        \mathcal{L}_{\text{SAR-W-MixMAE}} = \frac{1}{N} \sum_{n=1}^N W^n_{SAR} \left[ \left( \hat{t}_1^n - t_1^n \right)^2 (1 - m_n) + \left( \hat{t}_2^n - t_2^n \right)^2 m_n \right]    
    $}
\end{adjustbox}
\label{eq:loss_sar_w_mixmae}
\end{equation}

where $t_1^n, t_2^n$ represent the ground truth patches, $\hat{t}_1^n, \hat{t}_2^n$ are the reconstructed patches, $W^n_{SAR}$ is the pixel-wise weight for patch \(n\), and $m_n$ is the binary mask indicating visible patches. $(1 - m_n)$ focuses on patches from the first group, while $m_n$ focuses on patches from the second group.

SAR-W-MixMAE uses the Swin Transformer \cite{liu2021swin, liu2022swin} as its encoder, similar to MixMAE, but scaled with base configuration parameters: \(C = (128, 256, 512, 1024)\), \(H = (4, 8, 16, 32)\), and \(B = (2, 2, 18, 2)\), where \(C\), \(H\), and \(B\) denote the channel numbers, attention heads, and block numbers for each stage, respectively. Since the input resolution is reduced to \(128 \times 128\), the self-attention window sizes are adjusted to \((8, 8, 8, 4)\), ensuring compatibility with the smaller input size while maintaining effective attention operations.

Similar to MixMAE, we perform dual reconstruction by reconstructing masked tokens from the counterpart image. The pixel-weighted reconstruction training ensures that high signal power regions such as speckle regions contribute to the loss relatively less than other low power values areas on SAR images, improving robustness. The masking and mixing strategy is preserved, where the input consists of mixed representations of polarimetric SAR channels ($\text{VH}^\text{(dB)}$ and $\text{VV}^\text{(dB)}$) from two different sample images. This adaptation makes SAR-W-MixMAE a robust and efficient framework tailored specifically for SAR data.

\section{Experiments}
\label{sec:experiments}
\subsection{Dataset}
\textbf{BigEarthNetv1.0 \footnote{\url{https://bigearth.net/v1.0.html}}:} We utilized the large-scale publicly available BigEarthNet dataset \cite{sumbul2019bigearthnet} to pretrain our foundation model on Sentinel-1 Synthetic Aperture Radar (SAR) data. This dataset comprises 590,326 pairs of Sentinel-1 and Sentinel-2 high-resolution image patches, providing a substantial volume of diverse data for training deep learning models in remote sensing. 

The Sentinel-1 SAR images in BigEarthNet have a resolution of $120 \times 120$ pixels, corresponding to a spatial resolution of 10 meters on the ground. Each SAR image includes two channels, representing dual-polarization VH (vertical-horizontal) and VV (vertical-vertical). In our method, we preprocess the SAR data by stacking these channels and resizing the input into $128 \times 128$ patches, resulting in a $2 \times 128 \times 128$ input structure. This configuration enables the model to effectively capture the spatial and channel-based information necessary for downstream remote sensing tasks. We used version-1 of this dataset along with the provided training, validation, and test splits. 

\textbf{SEN12-FLOOD \footnote{\url{https://github.com/ClmRmb/SEN12-FLOOD}}:} We utilized the SEN12-FLOOD dataset \cite{rambour2020flood} for a downstream task of flood detection. We used the provided training and testing splits for SAR data. The dataset comprises 335 sequences \footnote{Although the dataset description states there are 336 sequences, the downloaded dataset contains a total of 335 sequences.}, with 68 sequences allocated for testing and the remaining sequences for training. On average, each sequence contains 14 SAR images with a resolution of $512 \times 512$ pixels and a 10-meter ground resolution. The authors of \cite{rambour2020flood} process the dataset sequentially using a GRU-based model with multiple inputs; however, further details of their approach are not clearly provided.

Since their approach was not applicable to our case, we adapted the dataset for pairwise input. Specifically, each pair consists of a non-flood image as the first image, while the second image could be either flooded or non-flooded. The dataset contains some partially covered earth surfaces, where certain examples include zero values for non-covered areas. These partially covered images can lead to erroneous pairs when matched with fully-covered images. To address this, we excluded partially covered images by disregarding instances (VV and VH) where at least 25\% of the pixel values are zero. After this curation, our dataset remains consistent with the original training and testing splits, resulting in 11,078 training pairs and 2,662 test pairs.

\subsection{Evaluation Metrics}
For BigEarthNet multi-label classification, we evaluate performance across different setups using multi-label Average Precision, F1 Score, and Precision for both macro and micro metrics. \textit{Macro} metrics calculate the average performance by giving equal weight to each class, regardless of class size, while \textit{Micro} metrics aggregate contributions from all classes, weighting results based on the total number of instances.

For SEN12-FLOOD binary classification in the flood detection downstream task, we assess the model's performance using Average Accuracy, Precision, Recall, and F1 Score.

\subsection{Pretraining Details}
We perform self-supervised pretraining on the BigEarthNet dataset using normalized 2-channel inputs (VH, VV) from SAR data, with normalization applied based on the calculated mean and standard deviation of the VH and VV bands. By default, the input size is set to $128 \times 128$.  

To maintain consistency with the underlying baseline model, we follow the pretraining hyperparameters of MixMAE \cite{liu2023mixmae}. The training of proposed SAR-W-MixMAE employs the AdamW optimizer \cite{loshchilov2017decoupled} with a learning rate of $1 \times 10^{-3}$ and a cyclic scheduler that decays the learning rate to 0. A warm-up phase of 40 epochs is applied, followed by training for a total of 64 epochs unless stated otherwise. In our experiments, we observed that longer epochs of pretraining did not have significant effect on downstream task performance except for the pretraining task on auto-encoder image reconstruction.

\subsection{Multi-Label Scene Classification on SAR Data of BigEarthNet}
In this task, the encoder of the pretrained model serves as the backbone, while a downstream task layer for multi-label classification is employed for predictions. The pretrained model is fine-tuned for 50 epochs, including a warmup phase of 5 epochs, using the multi-label soft margin loss with the AdamW optimizer. The training, validation, and testing sets are consistent with prior work \cite{sumbul2019bigearthnet}, comprising 269,695, 123,723, and 125,866 image samples, respectively.

Table \ref{table:bigearthnet_multiclass} presents the test results of our SAR-W-MixMAE method alongside the base MixMAE model \cite{liu2023mixmae}, where the inputs are adapted to the dataset (\textit{i.e.}, $2 \times 128 \times 128$), and classification results with random initialization, 
The test results demonstrate the effectiveness of the proposed model pretrained on the BigEarthNet dataset compared to a randomly initialized version, the baseline method.
The pretrained model shows a significant performance gain over the randomly initialized counterpart across all metrics, emphasizing the importance of the self-supervised pretraining for improving downstream task performance. For example, the Macro Average Precision increases from 0.6107 to 0.7088, and the Macro F1 score improves from 0.4936 to 0.6068. 

These results show how leveraging pretrained features enhances the model's ability to generalize, even when fine-tuning on a dataset with complex spatial and spectral patterns. Furthermore, our approach consistently slightly better than the MixMAE baseline model in almost all metrics, demonstrating that higher SAR weighting on low-scatter power regions did not have any adverse effect on general representation ability. 

\begin{table}[t!]
    \centering
    \caption{Test results on the BigEarthNetv1 dataset}
    
    \label{table:bigearthnet_multiclass}
    \renewcommand{\arraystretch}{1.5} 
    \begin{adjustbox}{width=\columnwidth}
        \begin{tabular}{l|c|cc|cc|cc}
            \hline
            \multirow{2}{*}{\textbf{Pretraining Method}} &
            \multirow{2}{*}{\textbf{Backbone}} &
            \multicolumn{2}{c|}{\textbf{Avg. Prec.}} &
            \multicolumn{2}{c|}{\textbf{F1}} &
            \multicolumn{2}{c}{\textbf{Precision}} \\
            \cline{3-8}
                   &
                   &
             Macro &
             Micro &
             Macro &
             Micro &
             Macro &
             Micro \\
            \hline
            Random Init &
            SwinB  &
            0.6107 & 
            0.7713 &  
            0.4936 & 
            0.6750 &  
            0.5484 & 
            0.7271  \\
            MixMAE (base model) \cite{liu2023mixmae} & 
            SwinB  &
            0.7044 & 
            0.8355 &
            0.6010 &
            0.7377 &
            0.6713 & 
            \textbf{0.7713}  \\
            \textbf{SAR-W-MixMAE (ours)} &
            SwinB           &
            \textbf{0.7088} &
            \textbf{0.8367} &
            \textbf{0.6068} &
            \textbf{0.7400} &
            \textbf{0.6748} &
            0.7692  \\
            \hline
        \end{tabular}
    \end{adjustbox}
\end{table}

\subsection{Flood Detection on SAR Data of SEN12-FLOOD}

For flood detection downstream task, we conduct experiments on the SEN12-FLOOD dataset \cite{rambour2020flood}. As described in the Dataset subsection, this dataset provides images with a resolution of $512 \times 512$ and binary classification labels: flooded or not flooded. We employ the pretrained model on BigEarthNet for this task. 

Since our foundation model, SAR-W-MixMAE, is designed for $128 \times 128$ dual-channel SAR image inputs, images from the SEN12-FLOOD dataset ($2 \times 512 \times 512$) are divided into $128 \times 128$ patches. Each sample image generates 16 encoder features, which are processed by the pretrained model. For a given pair, we compute the difference between the semantic features and apply average pooling to produce a single feature vector representation. The model is fine-tuned for 50 epochs using cross-entropy loss and the AdamW optimizer, with a warmup phase of 10 epochs.

Table \ref{tab:sen12flood} compares the proposed method's performance with the baseline MixMAE and a randomly initialized version. Both the proposed and baseline models are pretrained on BigEarthNet and fine-tuned on SEN12-FLOOD for binary flood/no-flood classification, while the randomly initialized version is trained on SEN12-FLOOD without pretraining.

The proposed model achieves the best performance across all metrics. Especially, pretraining on BigEarthNet significantly enhances recall (0.9027 vs 0.6903), emphasizing the importance of pretraining for enhancing model generalization in downstream tasks. The proposed model also performs significantly better than the baseline MixMAE in all metrics. This demonstrates the effectiveness of the weighting modification introduced in our SAR weighted ppretraining with SAR-W-MixMAE, which builds on MixMAE to achieve better overall performance.

\begin{table}[t!]
    \centering
    \caption{Performance comparison across Accuracy, Precision, Recall, and F1 Score. The proposed method improves the baseline MixMAE \cite{liu2023mixmae} and the randomly initialized version on the SEN12-FLOOD dataset \cite{rambour2020flood}.}
    \label{tab:sen12flood}
    \renewcommand{\arraystretch}{1.2} 
    \begin{adjustbox}{width=\columnwidth}
        \begin{tabular}{l|c|c|c|c|c}
            \toprule
            \textbf{Pretraining  Method} &
            \textbf{Backbone} &
            \textbf{Accuracy} &
            \textbf{Precision} &
            \textbf{Recall} &
            \textbf{F1} \\
            \midrule
            Random Init &
            SwinB  &
            0.8074 &
            0.7647 &
            0.6903 &
            0.7256 \\
            MixMAE (base model) \cite{liu2023mixmae} &
            SwinB  &
            0.8468 &
            0.7279 &
            0.8761 &
            0.7952 \\
            \textbf{SAR-W-MixMAE (ours)} &
            SwinB  &
            \textbf{0.8667} &
            \textbf{0.7727} &
            \textbf{0.9027} &
            \textbf{0.8327} \\
            \bottomrule
        \end{tabular}
    \end{adjustbox}
\end{table}

\section{Conclusion}
\label{sec:conc}

In this study, we explored the application of a masked auto-encoder model MixMAE \cite{liu2023mixmae} to Sentinel-1 SAR images and introduced SAR-W-MixMAE as a novel pretraining approach incorporating SAR-specific characteristics into the reconstruction loss. Proposed SAR-W-MixMAE addresses the challenges posed by SAR images by leveraging VH and VV polarimetric SAR channels in linear scale to compute pixel-wise weighting ($\text{W}_\text{SAR}$ in Fig.\ref{fig:method} and Eq.\ref{eq:weight_sar_w_mixmae}). Experiments on proposed pretrained model demonstrated promising improvements in downstream tasks on multi-label SAR image classification and flood detection, especially significant performance increase on flood detection downstream task. The experimental results on downstram tasks highlight the potential of foundation models designed based on SAR image characteristics. As a current limitation, it is important to explore the performance of SAR-W-MixMAE  with a speckle reduced SAR data, especially for the tasks requiring attention on the pixels with stronger backscatter power (brighter regions on SAR images such as buildings, man-made objects, etc. depending on the polarization type). Therefore, for that kind of specific task, it might be interesting to investigate weights computed to modulate the reconstruction loss for the brighter areas during MAE pretraining, especially for speckle noise filtered SAR data.

\section{Acknowledgement}
This work supported by AIST policy-based budget project ``R\&D on Generative AI Foundation Models for the Physical Domain".

\small
\bibliographystyle{IEEEtranN}
\bibliography{egbib}

\end{document}